\documentclass{article}
\usepackage{ijcai23}

\usepackage{times}
\usepackage{soul}
\usepackage{url}
\usepackage[hidelinks]{hyperref}
\usepackage[utf8]{inputenc}
\usepackage[small]{caption}
\usepackage{graphicx}
\usepackage{amsmath}
\usepackage{amsthm}
\usepackage{booktabs}
\usepackage{algorithm}
\usepackage{algorithmic}
\usepackage[switch]{lineno}
\usepackage{amsmath}
\usepackage{amssymb}

\title{RFENet: Towards Reciprocal Feature Evolution for Glass Segmentation}

\author{
Ke Fan$^1$
\and
Changan Wang$^2$\and
Yabiao Wang$^2$\and
Chengjie Wang$^{1,2}$\and
Ran Yi$^1$\footnotemark[1]\And
Lizhuang Ma$^1$\footnotemark[1]
\affiliations
$^1$Shanghai Jiao Tong University\\
$^2$Tencent Youtu Lab\\
\emails
slipperyfrank@sjtu.edu.cn,
\{changanwang, caseywang, jasoncjwang\}@tencent.com,
ranyi@sjtu.edu.cn,
ma-lz@cs.sjtu.edu.cn
}

\begin{document}

\maketitle

\renewcommand{\thefootnote}{\fnsymbol{footnote}} 
\footnotetext[1]{Corresponding authors.} 

\begin{abstract}
Glass-like objects are widespread in daily life but remain intractable to be segmented for most existing methods. The transparent property makes it difficult to be distinguished from background, while the tiny separation boundary further impedes the acquisition of their exact contour. In this paper, by revealing the key co-evolution demand of semantic and boundary learning, we propose a Selective Mutual Evolution (SME) module to enable the reciprocal feature learning between them. Then to exploit the global shape context, we propose a Structurally Attentive Refinement (SAR) module to conduct a fine-grained feature refinement for those ambiguous points around the boundary. Finally, to further utilize the multi-scale representation, we integrate the above two modules into a cascaded structure and then introduce a Reciprocal Feature Evolution Network (RFENet) for effective glass-like object segmentation. Extensive experiments demonstrate that our RFENet achieves state-of-the-art performance on three popular public datasets. Code is available at \it{\url{https://github.com/VankouF/RFENet}}.
\end{abstract}
\section{Introduction}
Detecting ubiquitous yet fragile glass-like objects is indispensable for vision based navigation systems. However, different from most other daily objects, glass-like objects are more confusing to be distinguished from background due to their 
transparent property. Besides, such objects mostly share an extremely thin separation boundary with background, making this task even more challenging.
\begin{figure}[!t]
\centering
\includegraphics[width=1\linewidth]{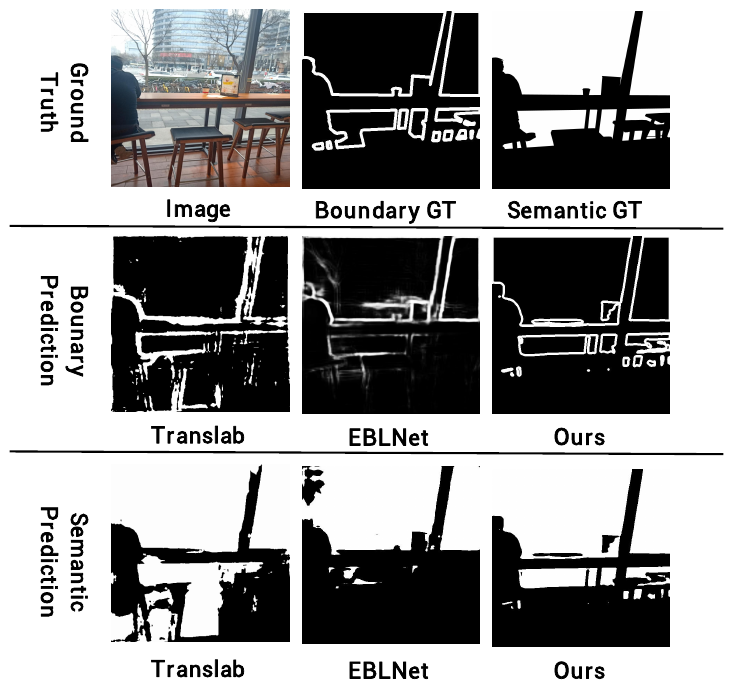}
\caption{Illustration of our predicted boundary and semantic results, compared with the two state-of-the-art methods in glass-like object segmentation, \textit{i.e.}, Translab and EBLNet. Our method produces a more accurate glass segmentation with a finer boundary, which is mainly contributed by our key feature co-evolution mechanism between semantic and boundary. The predictions of those ambiguous regions also gain the help from our SAR module. 
}
\label{comparison}
\end{figure}

Due to the above challenges, merely relying on either semantic content or separation boundary to segment out the glass regions 
is sub-optimal or at least inaccurate.
Although glass segmentation methods mainly rely on semantic features to predict the semantic map, adding boundary information can still help to obtain global shape context, thus improving the performance of segmentation. 
Meanwhile, boundary prediction could also be improved by auxiliary semantic assistance, which can help to reduce the false edge prediction and in turn facilitate a more accurate glass segmentation. 
Being consistent with the above observation, some previous glass segmentation works explored either assistance from boundary to semantic or assistance from semantic to boundary: 1) Using boundary to assist semantic: 
\cite{xie2020segmenting,zheng2022glassnet}  introduced an auxiliary boundary supervision as a guidance to conduct glass segmentation refinement, which helps the prediction of those uncertain regions around the boundary;  
2) Using semantic to assist boundary: \cite{he2021enhanced} proposed to supervise non-edge parts in a residual style to obtain finer edges and eliminate noisy edges in background, instead of directly performing a boundary feature enhancement. 
However, the above two paradigms, \textit{i.e.}, using boundary to assist semantic, or using semantic to assist boundary, both ignored the importance of feature co-evolution from the two sides 
(\textit{i.e.}, semantic branch and boundary branch).
In other words, previous methods did not conduct the bi-directional assistance between the 
two branches, 
and simply adopting one-way assistance results in inferior performance.

To address these issues, in this paper, we propose an adaptive mutual learning mechanism to enable the \textit{explicit} feature co-evolution between semantic branch and boundary branch. Such a mechanism helps to exploit the complementary information from each branch and is achieved by a novel Selective Mutual Evolution (SME) module. 
Specifically, the semantic feature is selectively enhanced with the guidance from the boundary branch, highlighting those weak response regions (especially for the pixels around boundary).  
In a similar way, the boundary feature is also selectively enhanced with the guidance from the semantic branch, 
mitigating the impact of edge noise from background or internal glass. 
With the above mutual learning strategy, the two branches reciprocally optimize each other to explore the intrinsic interdependence between them.

Despite the effectiveness of SME module, some regions remain indistinguishable, such as the pixels around boundary. To remedy this problem, we further propose a Structurally Attentive Refinement (SAR) module. To be more specific, we firstly sample a set of most reliable {\it boundary points} in the semantic feature, according to the confidence scores on the predicted boundary map, to capture the global shape information. Then a set of most {\it uncertain points} on glass segmentation map are enhanced with the semantic features of previous selected boundary points.
Notably, this adaptive enhancement process is conditioned on the contents of those uncertain points, and is achieved with a cross-attention operation.
In a nutshell, such an attentive feature refinement exploits extra boundary cues to help the inference of those ambiguous points, serving as a globally structural guidance. 
The proposed SAR module is pluggable with a simple design, and can also be applied to other boundary assisted methods.

Besides, inspired by pioneering exploration in the multi-scale feature representation, we further equip the above two modules with a cascaded style connection, benefiting from the progressive fusion of multiple receptive fields. 
Finally, we propose a Reciprocal Feature Evolution Network (RFENet) for glass-like object segmentation. We conduct extensive experiments against recent competitors on three popular glass-like object segmentation datasets and our RFENet achieves state-of-the-art performance. The visualized results in Figure \ref{comparison} also demonstrates the superiority of our RFENet. 

Overall, we summarize our contributions as follows:
\begin{itemize}
    \item 
    We propose RFENet, a novel glass-like object segmentation model, 
    which achieves state-of-the-art performance on three popular benchmarks.
    \item 
    We propose a Selective Mutual Evolution (SME) module to encourage the feature co-evolution of two branches, which effectively solves the inferior performance caused by only one-way assistance in the previous two-stream methods.
    \item 
    We propose a Structurally Attentive Refinement (SAR) module to conduct further feature refinement for uncertain points with useful global shape context. 
\end{itemize}

\section{Related Work}

\paragraph{Glass-like Object Segmentation.} It is much more challenging to segment out glass-like objects than those common objects, mainly due to that inner glass regions often share extremely confusing appearance with surrounding background. To remedy this problem, some methods \cite{xu2015transcut,chen2018tom,huo2022glass} resorted to exploit additional multi-modal information, such as 4D light-field, refractive flow map, and thermal image. Unfortunately, those multi-modal data is relatively expensive to acquire, which limits the wide applications. Instead, recent works \cite{yang2019my,mei2020don,lin2021rich,lin2020progressive,xie2020segmenting} contributed large-scale RGB image datasets for glass-like objects to promote research in related fields. However, due to the special property of glass-like objects, the off-the-shelf semantic segmentation methods\cite{chen2018encoder,zhao2017pyramid} failed to achieve a promising performance. Similarly, many state-of-the-art salient object detection approaches \cite{pang2020multi,qin2019basnet,zhuge2022salient,liu2022poolnet+} also result in an inferior prediction as the glass may not necessarily be salient.

Therefore, the demand for specialized methods recently attracts more attention in the field of glass-like object segmentation. \cite{yang2019my,mei2020don,lin2021rich} tried to integrate abundant contextual or contrasted features to help distinguish glass regions, implying the importance of contextual information. \cite{ji2023deep} utilizes a context and a texture encoder to extend the model from camouflaged object detection field into the glass detection area. Besides, \cite{xie2020segmenting,he2021enhanced,lin2020progressive} proposed to segment glass-like objects under the assistance of boundary cues, benefiting from the high localization accuracy of boundary. Inspired by existing research, we further reveal the importance of feature co-evolution demand for glass segmentation and boundary learning. Based on this observation, we propose an adaptive mutual learning mechanism to effectively exploit the complementary information between semantic and boundary.

\begin{figure*}[!t]
\centering
\includegraphics[width=0.9\linewidth]{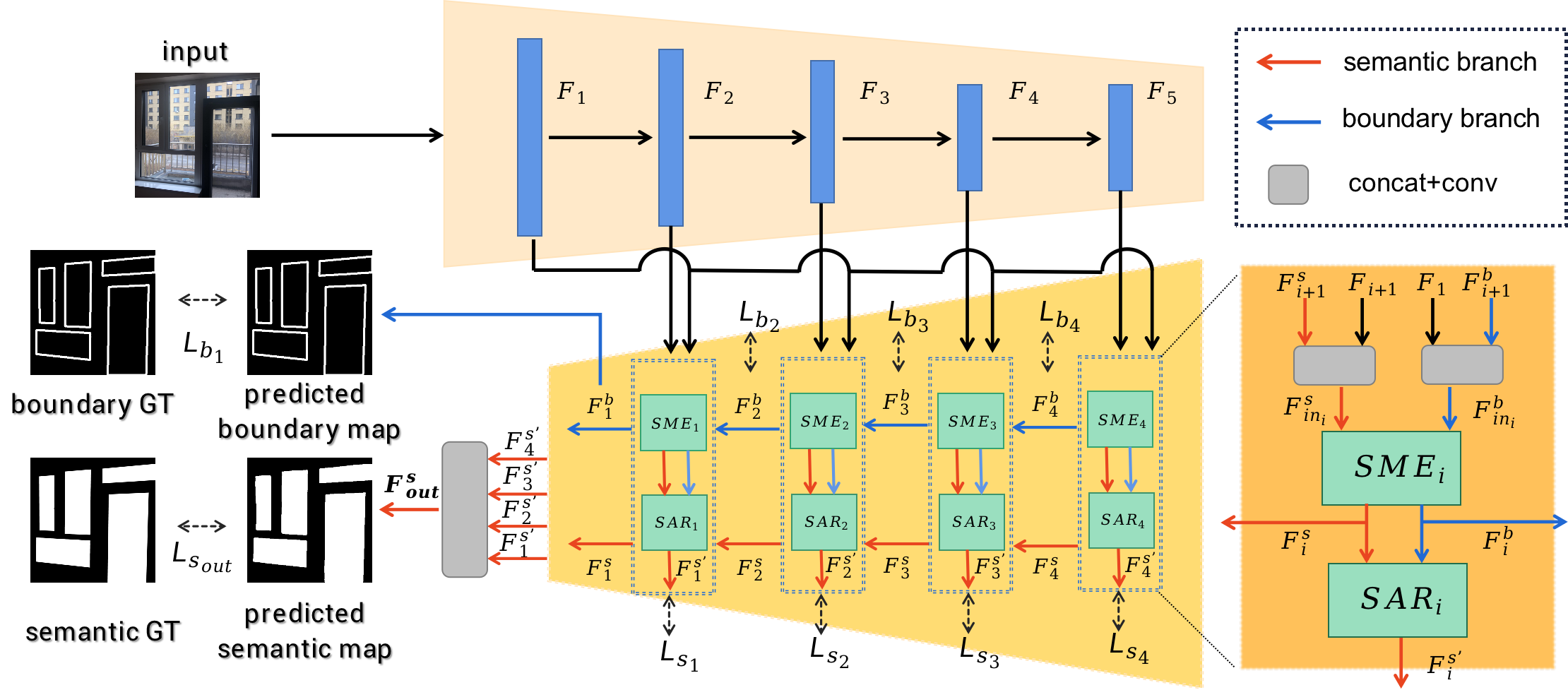}
\caption{
Overview of our proposed RFENet. The SME module enhances both semantic and boundary features in a reciprocal way, and the SAR module refines ambiguous points with global shape context. Both of them are combined repeatedly to form a cascaded structure.
}
\label{architecture}
\end{figure*}

\noindent\textbf{Boundary as Assistance}. The boundary contour of glass objects clearly defines their distribution range with a pixel-level localization ability. As a result, the effective exploration of boundary cues becomes crucial in high-precision glass segmentation. Actually, previous research has also demonstrated that introducing boundary cues into their model plays an important role for semantic segmentation \cite{takikawa2019gated,ding2019boundary} and salient object detection \cite{wang2019salient,zhao2019egnet,li2018contour}. As for glass-like object segmentation, \cite{xie2020segmenting,zheng2022glassnet} proposed to explicitly predict boundary map or decoupled boundary map, and utilized it as a guidance to assist the semantic stream. And \cite{he2021enhanced} tried to supervise edge part as well as none-edge part to explicitly model glass body and boundary. Both of them has proved the non-negligible performance gain brought by an appropriate exploit of boundary cues.

However, boundary prediction is often susceptible to background noise, especially for glass-like objects with transparent property. Different from existing methods, we propose an adaptive mutual learning mechanism to encourage feature co-evolution between semantic branch and boundary branch. Such a novel reciprocal structure helps to reduce the impact from background noise or internal glass reflection, with the guidance from semantic feature. Notably, the improved boundary prediction will in turns facilitate the glass segmentation in a cyclic enhancement style.

\noindent\textbf{Glass Segmentation Refinement.} The adoption of segmentation refinement technics has been demonstrated effective to further boost the model's prediction accuracy. For example, \cite{krahenbuhl2011efficient,he2010guided} proposed to use Conditional Random Fields and Guided Filter as post-processing to refine segmentation predictions. PointRend \cite{kirillov2020pointrend} and MagNet \cite{huynh2021progressive} proposed to refine some selected point set using local or global context information. In the field of glass-like object segmentation, EBLNet \cite{he2021enhanced} proposed a PGM module to exploit global shape prior, which further improves the edge prediction precision.

Differently, we propose to adaptively aggregate useful shape context for the most ambiguous points instead of certain boundary points. And the refined point set is dynamically sampled along with the optimization process, imitating hard sample mining strategy. The proposed refinement module provides structural context for predictions of some local regions, such as boundary and reflective regions.

\section{Method}
\subsection{Overview}
The architecture of our RFENet is illustrated in Figure \ref{architecture}. There are two parallel branches in our RFENet: semantic branch and boundary branch. Specifically, we firstly adopt 
ResNet50 \cite{he2016deep} with ASPP module \cite{chen2018encoder} as the backbone network to extract multi-scale deep features. Then Selective Mutual Evolution (SME) module is proposed to encourage the feature co-evolution of the two branches. Then we propose Structurally Attentive Refinement (SAR) module to further refine those uncertain points with global shape context. Finally, we integrate the two modules into a cascaded structure to exploit the internal hierarchical feature representation.

Formally, we denote the deep feature representations from different stages in backbone network as $F_i$, where $i \in \{1,2,3,4,5\}$ represents the $i$-th stage.  
The semantic branch predicts glass regions based on the last feature map $F_{in}^s$ (\textit{i.e.}, $F_5$), which contains the most rich context information. 
For the boundary map prediction, we use a concatenation\footnote{We perform bilinear interpolation on $F_5$ so that it has the same resolution as $F_1$, \textit{i.e.}, 1/4 of the original image size. We use the same operation for other features $F_i$ when necessary.} of $F_1$ and $F_5$ as input feature $F_{in}^b$ to take advantage of the texture information from low-level features. 
As shown in Figure \ref{architecture}, SME takes semantic feature and boundary feature as inputs, and obtain the co-evolved features by a mutual operation. Then the evolved features are both fed into SAR module to conduct a further refinement for semantic features under the guidance of boundary cues. The final prediction is obtained by a sequentially stacking of the two modules.

During each stacking process, for SME, we fuse lower-level features to recover the textural details for semantic branch, which gradually aggregates useful multi-scale context information. 
For boundary prediction, we repeatedly fuse the finest feature $F_1$ into the input feature of every stage to exploit more detailed texture information.  
The whole stacking process within SME module could be formulated as:

\begin{equation}
    {F_{i}^{s}}, {F_i^b} = \left\{\begin{matrix}
\mathrm{SME}_i \left(F_{i+1}, \left[F_{i+1};F_1\right]\right), & \mathrm{if}\;\;i = 4,\\ 
\mathrm{SME}_i \left(\left[{F_{i+1}^s};{F_{i+1}}\right], \left[{F_{i+1}^b};{F_1}\right]\right), & \mathrm{else}. 
\end{matrix}\right.
\label{eq:sme}
\end{equation}
where [$\cdot$] represents the feature concatenation operation, $F_{i}^{s}$ and $F_i^b$ represent the co-evolved semantic and boundary features output by SME.
Then $F_i^{s}$ and $F_i^b$ are input into the SAR module and obtain a refined semantic feature $F_i^{s'}$ under the guidance of $F_i^b$, which could be formulated as:
\begin{equation}
    F_i^{s'} = \mathrm{SAR}\left( F_i^s, F_i^b\right).
\label{eq:sar}
\end{equation}

The concatenation of $F_i^{s'}$, $ i\in \{1,2,3,4\}$ is used as the final semantic feature $F_{out}^s$, which is responsible for the prediction of final semantic map. Besides, to encourage the progressive evolution of intermediate semantic and boundary features $F_i^{s'}$ and $F_i^b$, we also attach additional prediction heads on them with the supervision from their individual ground truths.

\subsection{Selective Mutual Evolution (SME) Module}

We firstly introduce the key motivation of our reciprocal feature co-evolution mechanism. Compared with the segmentation of other daily objects, glass-like objects are more difficult to be distinguished from background regions, mainly due to their transparent property. 
One feasible workaround is trying to exploit useful boundary cues as assistance, which is also consistent with the human visual perception mechanism. 
In such a way, the semantic features around potential boundary will be enhanced and get more attention, which helps the model to capture the extent of glass region. 
However, the separation boundary between glass and surroundings is mostly too tiny to conduct an accurate prediction.  
Inspired by human's visual attention mechanism, 
semantic information of glass objects can be used to suppress false glass boundaries and highlight the features around real glass boundaries. Therefore, 
we propose to encourage feature co-evolution between boundary branch and semantic branch, simultaneously exploiting boundary cues to assist semantic features and exploiting semantic cues to assist boundary features.
In a short word, a more accurate boundary prediction produces a better glass segmentation, which in turns facilitates a more accurate boundary map, and vice versa. 

\begin{figure}[!t]
\centering
\includegraphics[width=1\linewidth]{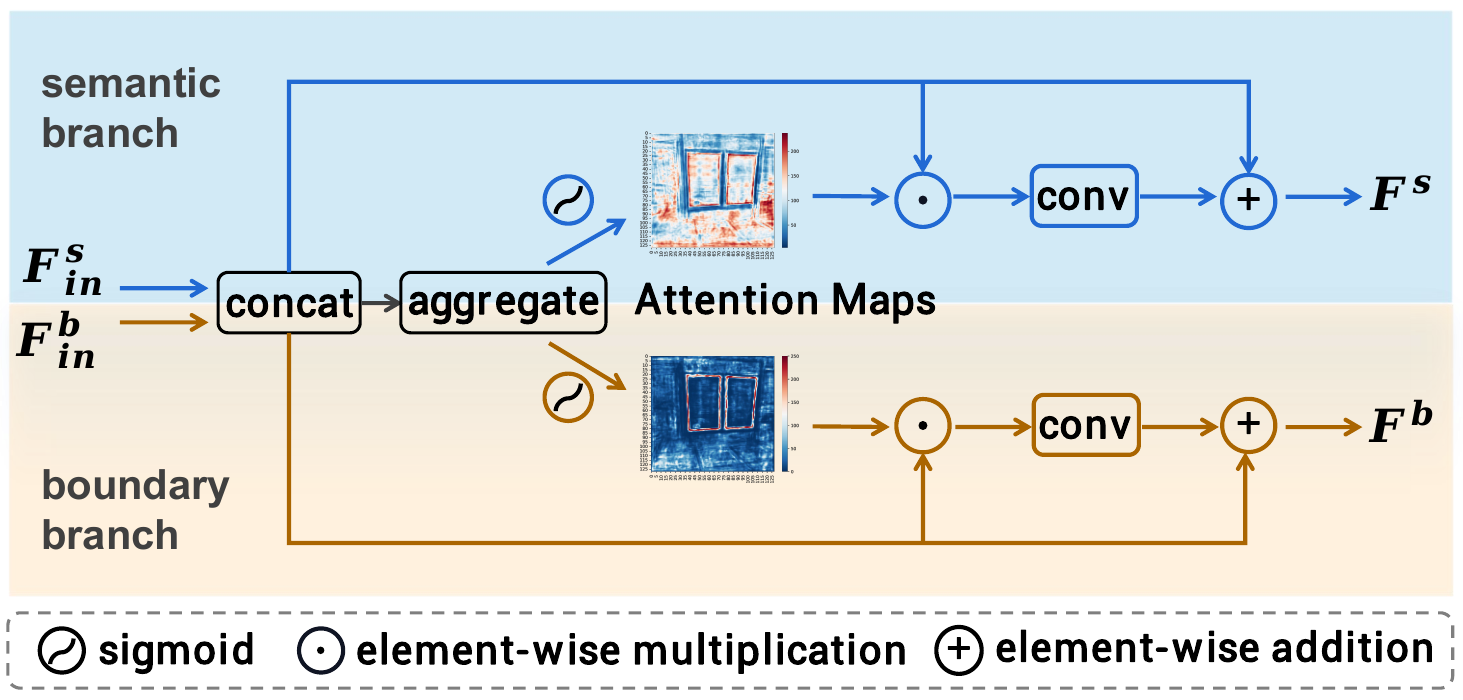}
\caption{
Illustration of the basic mutual block in our SME module. The attention maps are generated by a multi-branch aggregation operation on the semantic and boundary features, which helps to produce a more reliable attention score. 
Best viewed in color.
}
\label{sme}
\end{figure}

We then introduce the implementation details of our SME module. 
As shown in Figure \ref{sme}, each basic mutual block takes $F^s_{in}$ and $F^b_{in}$ as inputs and outputs corresponding features $F^s$ and $F^b$.
Each block generates a two-channel attention map $A$ from the joint feature representation of $F^s_{in}$ and $F^b_{in}$, and mutually enhances the input features using the attention map to capture complementary information from each other.

Specifically, the attention maps $A$ are generated with a multi-branch aggregation operation on the concatenated features $F_{in}^s$ and $F_{in}^b$.  The concatenated feature is fed into two branches. We first use a convolution with kernel size of $3$ to gather the local spatial context and then fuse the semantic and boundary information along the channel dimension to ensure a comprehensive view. Secondly, the fused feature is fed into two branches with convolutions of different kernel sizes (5 and 9). The large kernel based two-branch is designed to capture more long-range glass cues, which ensures a more reliable attention score. Finally, we stack several convolutions and a \textit{sigmoid} operation on the concatenation of above two features to predict the attention maps $A$. The above process can be formulated as:

\begin{equation}
    {A} ={\left[\begin{array}{c}
        {a^s} ; {a^b}
        \end{array}\right]}
        = {\sigma}(Aggregate([{F^s_{in}};{F^b_{in}}])),
\end{equation}
where $A \in {[0,1]}^{2 \times h \times w}$, $Aggregate$ represents the aggregation operation, and [$\cdot$] represents channel-wise feature concatenation. Then each channel of $A$ (\textit{i.e.}, $a^s$ and $a^b$) is used as the attention map to enhance $F^s_{in}$ and $F^b_{in}$ respectively in a residual manner:

\begin{equation}
    F^u = conv(F_{in}^u \bigodot a^u) + F_{in}^u, u \in \left \{s,b\right \},
\end{equation}
where $\bigodot$ denotes to element-wise multiplication. In this way, useful complementary information from either branch can be effectively exploited to adaptively enhance the features in the other branch without losing the original content.

We visualize the predicted two attention maps in Figure \ref{sme} to provide an intuitive explanation of our SME module. 
As shown in the figure, for the semantic branch, the features around the boundary are highlighted to accurately determine the 
distribution range of glass area.
Notably, some regions outside the glass also receive relatively high attention. We assume that the network could attentively focus on some useful contextual information.
As for the boundary branch, irrelevant contours in background regions are suppressed by relatively weak attention score to mitigate the noise disturbance, which is also in line with our analysis.

\subsection{Structurally Attentive Refinement (SAR) Module}
Despite the effectiveness of our SME module, there are still some difficult pixels remaining confusing to be distinguished, such as pixels located at boundaries, reflective regions and background regions with smooth surface. 
The inference of those ambiguous and difficult points is challenging without extra context priors. 
Fortunately, for any given difficult point, there are still several useful context cues to assist its inference, such as distance from its nearest boundary, curvature of local boundary and the whole shape of glass, \textit{etc.}. 
Therefore, inspired by human's visual reasoning mechanism, we propose to adaptively aggregate the glass shape context as structural priors to help the inference of those ambiguous points. 
We firstly sample the features located at the top-confidence {\it boundary points} to construct a feature set as a compact representation of the glass shape context. 
Then we dynamically select the most {\it uncertain points} to conduct feature refinement along with the learning process, which acts as a kind of hard sample mining strategy. 
The refinement process is conditioned on the content around the current uncertain points, and is guided by the feature similarity with those boundary points. 
With such a refinement design, those uncertain points are allowed to freely explore useful shape context cues without the constraint from limited receptive field.

\begin{figure}[!t]
\centering
\includegraphics[width=1\linewidth]{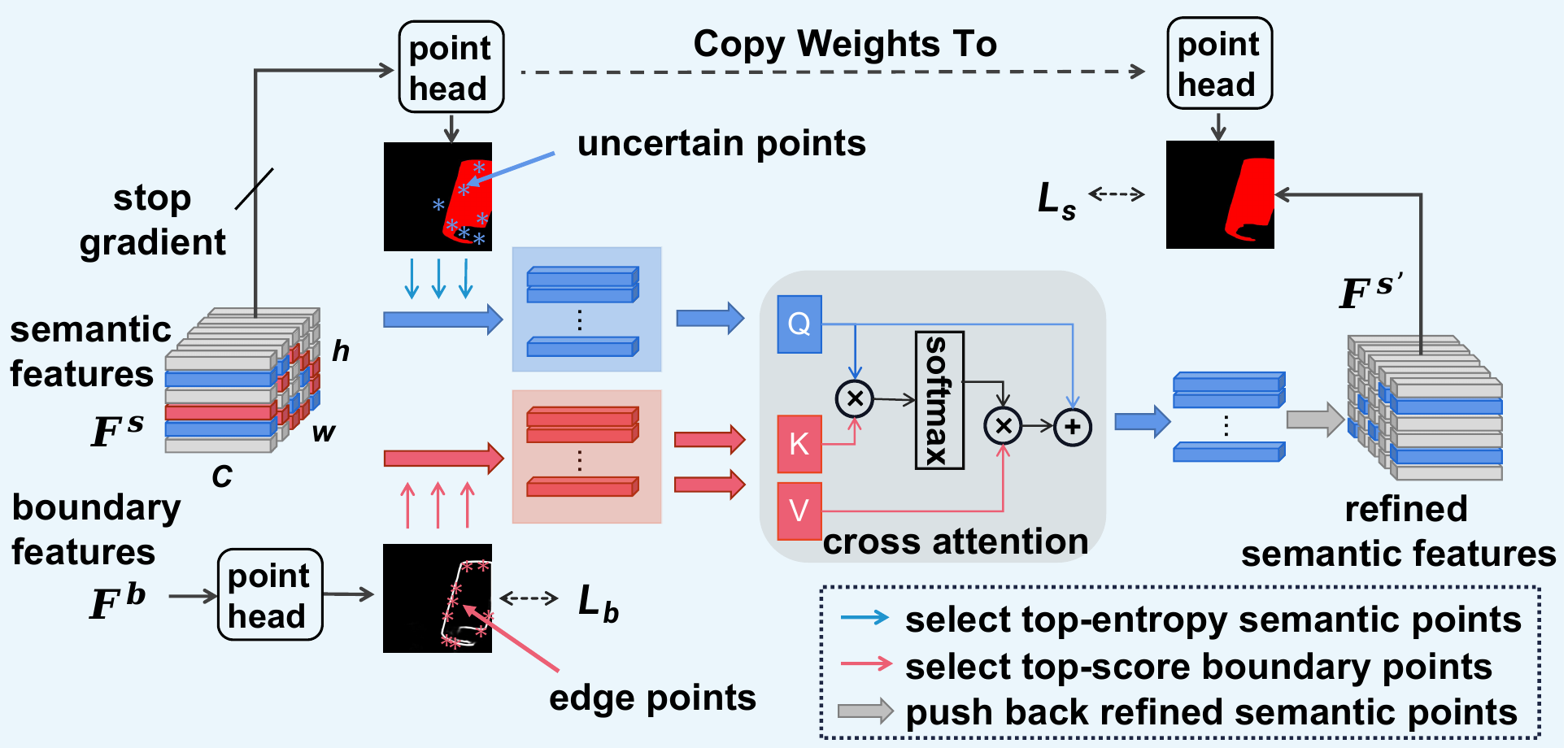}
\caption{
The illustration of our SAR module. The features of uncertain points are refined by global shape context.
}
\label{sar}
\end{figure}

Following the above methodology, we propose a Structurally Attentive Refinement (SAR) module, as illustrated in Figure \ref{sar}. 
The SAR module accepts the semantic feature $F^{s}$ and boundary feature $F^b$ from SME module as inputs, and outputs refined semantic feature $F^{s'}$. Based on the original input semantic features $F^{s}$, we can obtain initial glass prediction $P_s \in \mathbb{R}^{n\times h \times w}$ with $n$-class glass prediction head, and initial boundary map $P_b \in \mathbb{R}^{1\times h \times w}$ with edge prediction head. 
We use the prediction scores in $P_s$ and $P_b$ to select those uncertain points and top-confidence boundary points.  
Then we propose attentive feature refinement to enhance features of uncertain points under the assistance of top-confidence boundary points.
After the features of those uncertain points are enhanced, we push them back into $F^{s}$ according to their original indices to get the refined feature $F^{s'}$. 

{\bf{Attentive feature refinement.}} we introduce the details of the core attentive refinement operation as follows. 
1) Firstly, we iterate over all pixels in $P_s$ to calculate their Entropy for the measurement of prediction uncertainty. Then we select $K$ points with the top entropy as the most ambiguous points set, the semantic features $Q$ of which are later refined with extra shape context. 
2) Secondly, we select $M$ boundary points with the top prediction confidence based on $P_b$ to construct the boundary feature set $V$. Intuitively, the feature set $V$ could be considered as a compact representation of geometric information of glass, providing useful shape context.  
3) Finally, for each feature in uncertain semantic feature set $Q$, we adaptively aggregate the most relevant shape context feature from boundary feature set $V$ based on the feature correlation. 
We then obtain the refined features by fusing the aggregated results in a residual manner, which are later used for final semantic prediction. 
Notably, the above mentioned content-aware attentive refinement process can be easily implemented through the off-the-shelf cross-attention module, if we use the semantic features $Q$ as queries and use boundary features $V$ as both keys and values (\textit{i.e.}, $K$=$V$). There are usually several parallel attention heads in one cross-attention module, and each of which could be formulated as following:

\begin{equation}
    Attention(q,k,v) = softmax(\frac{q*k^t}{\sqrt{d_k}})\cdot v,
\end{equation}
where $d_k$ is the dimension of input features $q$, $k$ and $v$.

\begin{figure}[!t]
\centering
\includegraphics[width=1\linewidth]{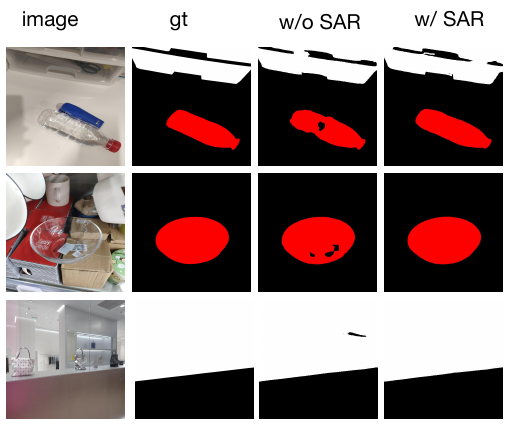}
\caption{
Visualized exhibition for the effectiveness of SAR module. The first and second columns are original images and their ground truths, and the third and fourth columns are the predicted maps without or with SAR module. Those ambiguous points, such as boundary or reflective points, can be refined with SAR module.
}
\label{uncertain_visualization}
\end{figure}

\subsection{The Cascaded Connection}
To fully explore the multi-scale feature representation within the backbone, we integrate our SME and SAR module into a cascaded structure. We denote the semantic and boundary features output from SME and SAR in stage $i$ to $F_i^s$, $F_i^{s'}$ and $F_i^b$, respectively. As shown in Equation \ref{eq:sme} and \ref{eq:sar}, we sequentially stack the two modules from higher level to lower level stage, making a cascaded optimization style. Besides, to aggregate more detailed information, $F_1$ from backbone is always concatenated with $F_{i+1}^b$ before fed into SME module at each stage $i$. The semantic features $F_i^{s'}$ output from every stage are concatenated to produce the final semantic map. At the same time, the intermediate predictions of each stage are also supervised to encourage a progressive feature evolution.

\subsection{Loss Design}
Our RFENet is supervised with a joint function of semantic loss and boundary loss, which can be formulated as:
\begin{equation}
    L =  L_{s_{out}} + \lambda_s L_{s_i} + \lambda_b L_{b_i}, i\in \{1,2,3,4\},
\end{equation}

where the semantic losses $L_{s_{out}}$ and $L_{s_i}$ are Cross-Entropy Losses and supervise the predictions from both the final glass prediction head, and the intermediate predictions of each stage in semantic branch. Meanwhile, the boundary losses $L_{b_i}$ supervise the intermediate predictions of each stage in boundary branch. Considering that the boundary points only account for a small range in an image, we use the Dice Loss \cite{milletari2016v} as the boundary loss $L_{b_i}$ to avoid the sampling imbalance problem. We generate the ground-truth boundary maps following \cite{xie2020segmenting} with the thickness of 8. The $\lambda_s$ and $\lambda_b$ are used to balance the effect from $L_{s}$ and $L_{b}$, which are set to 0.01 and 0.25 for all experiments.


\section{Experiment}

\subsection{Datasets}
{\bf{Glass Datasets: }}\textbf{(1) Trans10k} \cite{xie2020segmenting} is a large-scale transparent object segmentation dataset, consisting of 10,428 images with three categories: things, stuff and background. Images are divided into 5,000, 1,000 and 4,428 images for training, validation and test, respectively. It is by far the largest transparent object segmentation dataset with the most detailed annotations. Taking into consideration of the data amount and scene diversity, we conduct most of our experiments on this dataset to ensure a convincing result.
\textbf{(2) GSD} \cite{lin2021rich} is a medium-scale glass segmentation dataset containing 4,098 glass images, covering a diversity of indoor and outdoor scenes. All the data are randomly split into a training set with 3,285 images and a test set with 813 images.  
We use GSD to validate the generalization ability of our method.

{\bf{Mirror Dataset:}}
\textbf{PMD} \cite{lin2020progressive} is a large-scale mirror dataset contains 5,096 training images and 571 test images. It contains a variety of real-world images that cover diverse scenes and common objects, making it much closer to practical application. We conduct experiments on PMD dataset to demonstrate our model's transferability for mirror segmentation although it is designed for glass-like objects.

\subsection{Evaluation Metrics}
We follow the previous works to mainly adopt the following metrics to evaluate the performance of our model: mean Intersection over Union (mIoU), Pixel Accuracy (Acc), Mean Absolute Error (mAE), mean Balance Error Rate (mBER) and F-score. The mIoU is widely used  
to calculate the ratio of true positive prediction. The mBer measures a more comprehensive error rate by taking the sample imbalance problem into consideration. The Acc is used to provide a rough estimation of the pixel-level classification ability. And the mAE provides a measurement for the absolute prediction error of the segmentation map. Besides, we also follow \cite{yang2019my,mei2020don} to measure F-score in PMD and GSD benchmark, which gives a more comprehensive view of Precision and Recall rate. 

\subsection{Implementation Details}
We implement RFENet using the PyTorch framework \cite{paszke2019pytorch}. The backbone network is initialized with ImageNet pre-trained weight, while the remaining parts are randomly initialized. ResNet50 is used for Trans10k and ResNeXt101 is used for GSD and PMD. We launch the training process on 4 GPUs with synchronized batch normalization, unless otherwise mentioned. For simplicity, we use stochastic gradient descent (SGD) as optimizer, which is scheduled with a poly policy with a power of 0.9.

For Trans10k dataset, input images are resized to a size of 512$\times$512 for both training and testing. The initial learning rate is set to 0.04, and weight decay is set to 0.0001. We use a mini-batch size of 4 for each GPU and run for 60 epochs.

For GSD dataset, following the same setting as \cite{lin2021rich}, the input images are firstly resized to 400$\times$400 and then randomly cropped to 384$\times$384.
Random flipping is used for training. 
During inference, the test images are also first resized to 384 × 384 before fed into the network. 
The initial learning rate is set to 0.01, 
and weight decay is set to 0.0005. We run for 80 epochs with a batch size of 6 for each GPU.

For PMD dataset, we adopt the same setting as PMDNet \cite{lin2020progressive}, where the input images are resized to 384$\times$384. The initial learning rate is set to 0.03. The other settings remain the same as those on GSD dataset. 

\begin{table}[]
\resizebox{\columnwidth}{!}{%
\begin{tabular}{l|c|c|c|c}
\hline
Model      & mIoU↑           & Acc↑            & mAE↓            & mBER↓          \\ \hline
BiSeNet \small \shortcite{yu2018bisenet}    & 73.93          & 77.92          & 0.140          & 13.96         \\
DenseAspp \small \shortcite{yang2018denseaspp} & 78.11          & 81.22          & 0.114          & 12.19         \\
DeeplabV3+ \small \shortcite{chen2018encoder} & 84.54          & 89.54          & 0.081          & 7.78          \\
FCN    \small \shortcite{long2015fully}    & 79.67          & 83.79          & 0.108          & 10.33         \\
PSPNet  \small\shortcite{zhao2017pyramid}   & 82.38          & 86.25          & 0.093          & 9.72          \\
Translab  \small\shortcite{xie2020segmenting} & 87.63          & 92.69          & 0.063          & 5.46          \\
EBLNet(OS16) \small\shortcite{he2021enhanced}    & 89.58          & 93.95          & 0.052          & 4.60           \\
EBLNet(OS8) \small\shortcite{he2021enhanced}    & 90.28          & 94.71          & 0.048          & 4.14          \\ \hline
Ours(OS16)       & \textbf{90.97} & \textbf{96.32} & \textbf{0.044} & \textbf{3.82} \\
Ours(OS8)       & \textbf{91.25} & \textbf{96.50} & \textbf{0.043} & \textbf{3.68} \\ \hline
\end{tabular}
}
\caption{Quantitative comparison between the proposed RFENet and state-of-the-art methods on Trans10k test set. OS means the output stride in the backbone network.}
\label{tab_trans_sota}
\end{table}
    \subsection{Comparison With the State-of-the-Arts}
To demonstrate the superiority of our method, 
we conduct extensive experiments on three datasets, where we compare with recent state-of-the-art methods of glass-like object segmentation as well as some representative methods in common objects semantic segmentation.
The semantic segmentation candidates to be compared are selected by referring to \cite{he2021enhanced}.
 
\textbf{Quantitative Evaluation.} Firstly, as shown in Table \ref{tab_trans_sota}, off-the-shelf methods that are designed for common objects produce inferior performance, such as the best one DeeplabV3+(ResNet50). This is in line with our analysis that the special transparent property of glass-like objects make it challenging to directly segment out without extra assistance. 

Secondly, we make a comparison with the two recent strong competitors that are also designed for glass-like objects, \textit{i.e.}, Translab \cite{xie2020segmenting} and EBLNet \cite{he2021enhanced}. As shown from Table \ref{tab_trans_sota}, our RFENet with the typical output stride of 16 achieves an impressive improvement with at least 1.5\% gain in mIoU. Besides, when we adopt a finer feature map resolution of stride 8, our RFENet sets new records for all metrics on the Trans10k dataset. This promising improvement demonstrates the effectiveness of our core claim that encouraging the feature co-evolution between semantic branch and boundary branch helps to maximize the exploitation of their complementary information.

Thirdly, as shown in Table \ref{tab_gsd_sota}, on the GSD dataset, our RFENet achieves an improvement of 3.4\% in terms of mIoU, compared with previous SOTA method GlassNet \cite{lin2021rich}. The consistent improvement and competitive performance on GSD dataset demonstrates the generalization ability of our RFENet on other datasets, which is crucial for practical applications. 
We further extend our glass segmentation method to a mirror dataset, the PMD dataset.
As shown in Table \ref{tab_pmd_sota}, our RFENet achieves even more improvement on mIoU, which clearly demonstrates the transferability of our method. 

\textbf{Qualitative Evaluation.} Our method also achieves superior qualitative results which we exhibit in the appendix.

\begin{table}[]
\resizebox{\columnwidth}{!}{%
\begin{tabular}{l|l|c|l|l}
\hline
Model    & mIoU↑         & $F_\beta$↑              & mAE↓            & mBER↓          \\ \hline
GDNet \small\shortcite{mei2020don}     & 79.01         & 0.869          & 0.069          & 7.72          \\
Translab \small\shortcite{xie2020segmenting} & 74.05                              & 0.837          & 0.088                               & 11.35                             \\
GSD \small\shortcite{lin2021rich}      & 83.64                              & 0.903          & 0.055                               & \textbf{6.12}                              \\
PGSNet \small\shortcite{yu2022progressive}  & 83.65                              & 0.868          & 0.054                               & 6.25                              \\ \hline
Ours     & \textbf{86.50} & \textbf{0.931} & \textbf{0.048} & 6.23 \\ \hline
\end{tabular}
}
\caption{Quantitative comparison between the proposed RFENet and state-of-the-art methods on GSD test set.}
\label{tab_gsd_sota}
\end{table}
\begin{table}[]
\begin{center}
\begin{tabular}{l|l|c}
\hline
Model       & \multicolumn{1}{c|}{mIoU↑}       & $F_\beta$↑          \\ \hline
MirrorNet \small\shortcite{yang2019my}   & \multicolumn{1}{c|}{58.51}      & 0.748      \\
PMDNet \small\shortcite{lin2020progressive}      & 66.05                           & 0.792      \\
VCNet \small\shortcite{tan2022mirror}      & 68.25                           & 0.812      \\
Guan \textit{et al.} \small\shortcite{guan2022learning} & 66.84                           & 0.844      \\ \hline
Ours        & \textbf{73.56}                  & \textbf{0.851} \\ \hline
\end{tabular}
\caption{
Quantitative result of extending our RFENet to mirror dataset PMD and comparison with state-of-the-art mirror segmentation methods.}
\label{tab_pmd_sota}
\end{center}
\end{table}

\subsection{Ablation Study}
In this section, we conduct extensive ablation study to demonstrate the effectiveness of SME and SAR modules. All experiments are conducted on a single GPU for efficiency.

\noindent\textbf{Effectiveness of SME module}. We use a two-stream network as our baseline method, in which we directly attach the glass prediction head and edge prediction head to the backbone feature $F_{in}^s$ and $F_{in}^b$. As shown in Table \ref{tab_ablation}, we firstly add SME$_4$ to conduct a single-scale mutual learning, which achieves a significant improvement with 1.5\% in mIoU. 
 
Notably, similar improvement from the SME module can also be achieved even if we have added the SAR module, which implies that the two proposed modules work in a complementary way. These quantitative results strongly demonstrate the effectiveness of 
the feature co-evolution between semantic feature and boundary feature.

\begin{table}[]
\resizebox{\columnwidth}{!}{%
\begin{tabular}{lllcccc}
\hline
SME & SAR & Cascade & mIoU↑           & Acc↑           & mAE↓            & mBER↓          \\ \hline
    &     &         & 88.37          & 95.22         & 0.057          & 4.89          \\
\checkmark   &     &         & 89.70          & 95.56         & 0.050          & 4.33          \\
\checkmark   &     & \checkmark       & 90.07          & 95.86         & 0.048          & 4.15          \\
    & \checkmark   &         & 88.66           & 95.27         & 0.056          & 4.69          \\
\checkmark   & \checkmark   &         & 90.04          & 96.07         & 0.049          & 4.02          \\
\checkmark   & \checkmark   & \checkmark       & \textbf{90.42} & \textbf{96.10} & \textbf{0.046} & \textbf{3.99} \\ \hline
\end{tabular}
}
\caption{The effects of SME and SRA module. We use a naive two-stream (\textit{i.e.}, semantic and boundary) network as our baseline method.}
\label{tab_ablation}
\end{table}
\begin{table}[]
\resizebox{\columnwidth}{!}{%
\begin{tabular}{cccccc}
\hline

\begin{tabular}[c]{@{}c@{}}semantic\\ attention\end{tabular} & \begin{tabular}[c]{@{}c@{}}boundary\\ attention\end{tabular} & mIoU↑ & Acc↑ & mAE↓ & mBER↓  \\ \hline
   &                                                              & 88.37                     & 95.22                     & 0.057                     & 4.89                     \\
 \checkmark  &                                          & 89.41 & 95.41 & 0.052 & 4.34 \\
   &  \checkmark                                       & 88.65 & 94.40 & 0.056 & 4.69 \\
\checkmark                          & \checkmark                                                            & \textbf{89.70}                     & \textbf{95.56}                     & \textbf{0.050}                     & \textbf{4.33}                     \\ \hline
\end{tabular}
}
\caption{The analysis of mutual learning mechanism. Both the two one-way assistance result in inferior performance.}
\label{single_mutual}
\end{table}

For a more in-depth analysis on the mutual learning mechanism, we implement the one-way assistance by replacing the attentive feature enhancement operation in our SME module with an identity connection. To avoid any potential information flow from the other side, we also stop the gradient back-propagation from the generated attention map. As shown in Table \ref{single_mutual}, both the two one-way assistance strategies produce inferior performance, compared with the bi-directional assistance. It is worth noting that the combination of the two one-way assistance achieves much more improvement than the summation of their individual improvement, which shows the benefit of the feature co-evolution.

For further qualitative analysis of the attention map, we illustrate it in the appendix.

\noindent\textbf{Effectiveness of SAR module.} As shown in Table \ref{tab_ablation}, the SAR module can achieve a further improvement in all metrics on the basis of the SME module. Similar improvement can also be consistently observed for the two-stream baseline model. Since the SME module only provides a local feature enhancement, these results effectively demonstrate the indispensability of assistance from global shape context. Qualitatively, from Figure \ref{uncertain_visualization} we can clearly see that without the SAR module to select and refine uncertain points, there are indeed some points that are difficult to predict correctly.

\noindent\textbf{Effectiveness of cascaded connection.} As shown in Table \ref{tab_ablation}, the cascaded connection achieves consistent improvement. It implies the indispensability of multi-scale representation.

\section{Conclusion}

In this paper, we tackle the challenging problem of glass-like object
segmentation with our proposed RFENet. The model contains two novel modules, Selective Mutual Evolution module for reciprocal feature learning between semantic and boundary branch, and Structurally Attentive Refinement module for refining those ambiguous difficult points through the global shape prior. Extensive experiments show that our model achieves state-of-the-art performance on Trans10k, GSD and PMD datasets.

\appendix
\section{Appendix}
\subsection{Overview}
In this appendix, more visualization results and analysis, as well as the additional details are provided, which are organized as follows:
\begin{itemize}
    \item Sec.~\ref{sec:attention_map} provides the visualization of attention maps by heatmap technique and the analysis of how double branches interact with each other via the attention mechanism.
    \item Sec.~\ref{sec:qualitative} provides qualitative comparisons on Trans10k \cite{xie2020segmenting} (Sec.~\ref{ssec:trans10k}) and PMD \cite{lin2020progressive} (Sec.~\ref{ssec:pmd}), where our method achieves state-of-the-art results.
    \item Sec.~\ref{sec:additional details} provides the additional details analysis, including inference time comparison (Sec.~\ref{ssec:inference time}) with current state-of-the-art method EBLNet \cite{he2021enhanced} and stability analysis (Sec.~\ref{ssec:stability analysis}).
\end{itemize}

\subsection{Attention Map Visualization}
\label{sec:attention_map}
\begin{figure*}[!t]
\centering
\includegraphics[scale=0.6]{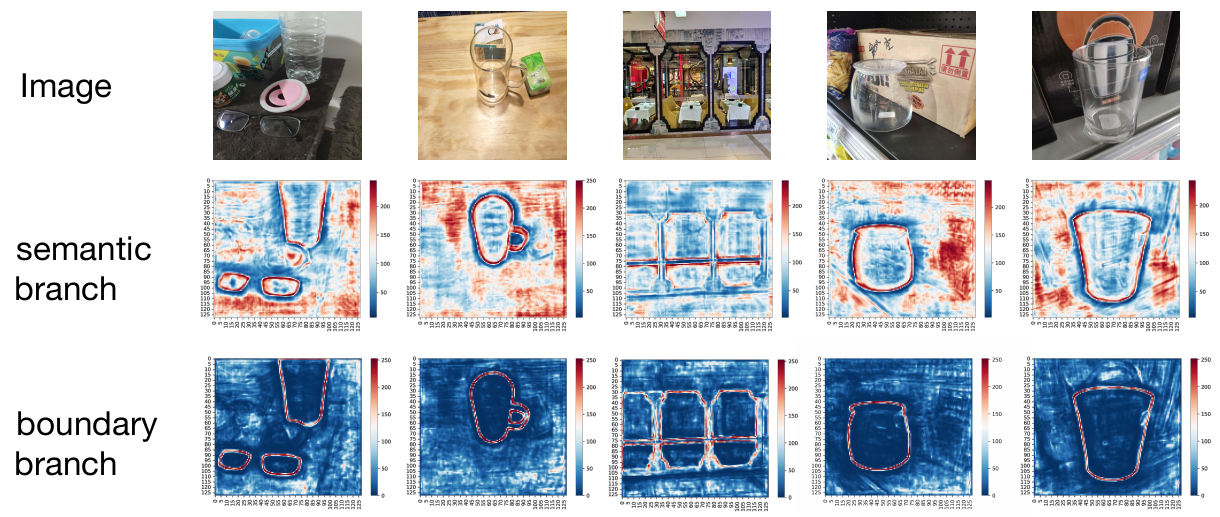}
\caption{Visualization of predicted semantic and boundary attention maps in our SME module. These results intuitively illustrate the way that SME utilizes mutual information between semantic and boundary features.
}
\label{singlelayer-attention}
\end{figure*}

Our SME module encourages the co-evolution between semantic and boundary features by predicting corresponding attention maps. The semantic information conveyed to the boundary branch helps to suppress false boundaries and highlight the features around real boundaries. Boundary branch features help the semantic branch to determine a more accurate object’s distribution range. To analyze how the two branches interact with each other by attention mechanism, we visualize the attention by heatmap technique. Firstly, we visualize the attention of the model without cascade in Figure \ref{singlelayer-attention}, where a more saturated color in red means a higher score.

As shown in Figure \ref{singlelayer-attention}, for the semantic branch, the features around the boundary are highlighted to locate the distribution extent of the glass area precisely. 
Notably, some outer glass regions also have high attention score, which may indicate that the model could selectively focus on some helpful context.

For the boundary branch, the irrelevant contours in background regions are suppressed by a relatively weak attention score. It demonstrates that semantic information can mitigate the impact of false edges in the background, which is in line with human's visual attention mechanism. 

\begin{figure*}[!t]
\centering
\includegraphics{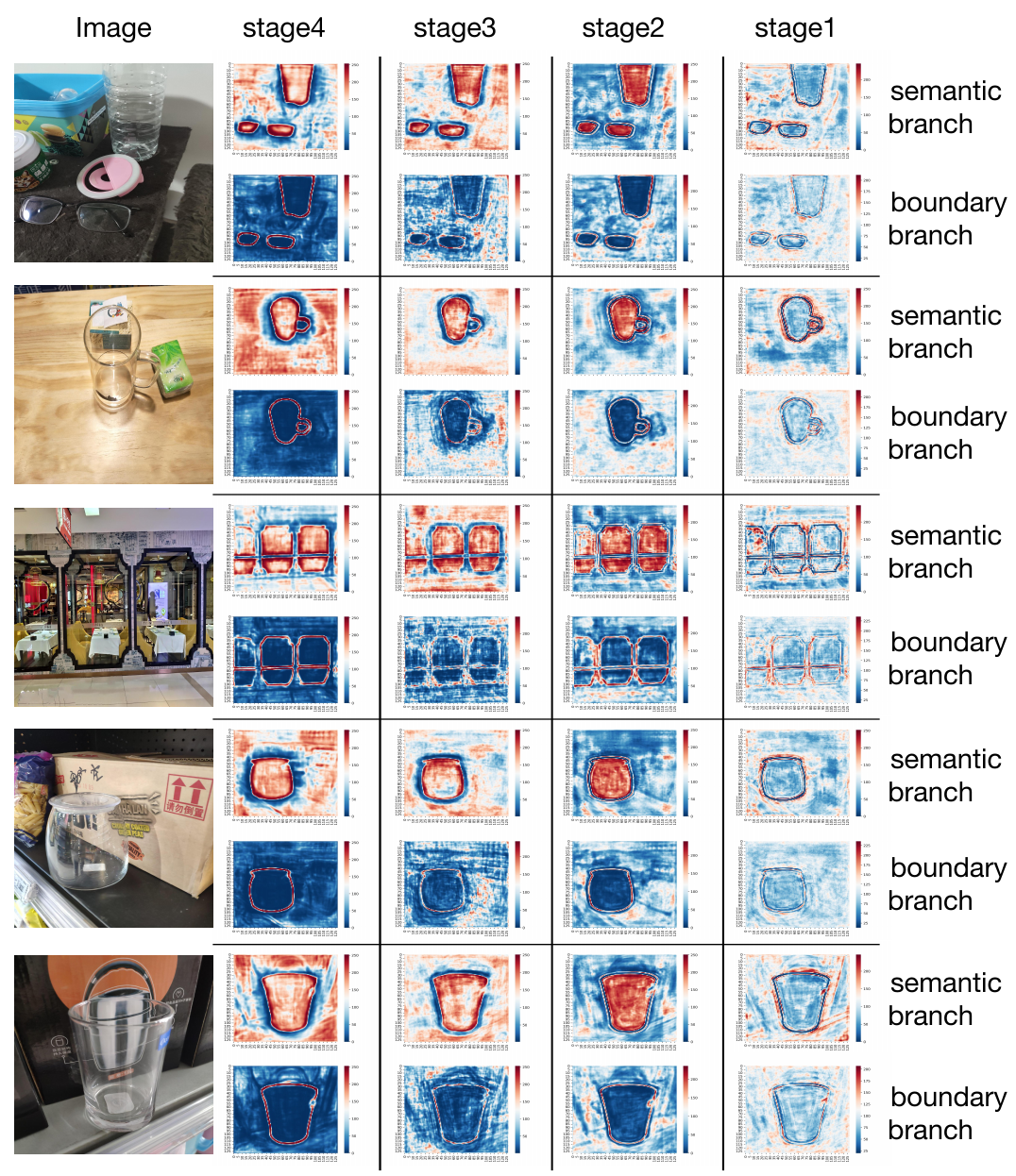}
\caption{Visualization of predicted semantic and boundary attention maps at each stage in our SME module with cascade structure. In each grid, the attention in the first row represents the semantic branch attention, and the other in the second row represents the boundary branch attention.
}
\label{cascade-attention}
\end{figure*}

Furthermore, we visualize attention at each stage by adding the cascade structure, where we explore how multi-scale operation promotes our mutual evolution. The images and their order we visualized are exactly the same as Figure \ref{singlelayer-attention}. As demonstrated in Figure \ref{cascade-attention}, for the semantic branch, as the stage goes deeper (index $i$ gets smaller), we find attention progressively moves closer to the object boundary. The attention at the initial two stages (stage4, stage3) tends to focus on both the inner and outer region of the glass. The attention on the outside of the glass region gets weaker at the penultimate stage (stage2), but still pays much attention to the inner region. The attention of the final stage further weakens the importance of the inner region, which implies the model is striving for overcoming the texture variety problem caused by the transparency attribute.  We analyze that it owes to the constraint information received from the boundary which gets more accurate as the stage goes deeper.

As for the boundary branch, attention at each stage attaches the most importance to the contour of the target, instead of other irrelevant boundaries, which is in line with our assumption that the semantic information conveyed to the boundary branch could mitigate the noise disturbance. Besides, we also notice that as the stage goes deeper, the attention (red) on the boundary gets thinner, and the contrast of attached importance between the boundary and other region gets lower. We assume it is probably because the boundary feature gradually becomes accurate, and it is not necessary to consistently attach quite different attention against the boundary and other regions. The model fits this phenomenon dynamically during the whole cascade process. 

Besides, the overall process demonstrates that the cascaded structure could promote the model performance in a coarse-to-fine manner, which is indispensable for network discrimination.

In conclusion, our SME module effectively utilizes complementary information from the boundary and semantic features by conveying information with attention maps.

\subsection{Qualitative Evaluation}
\label{sec:qualitative}
Our RFENet model achieves superior results compared with other methods. We exhibit the visualization results on Trans10k \cite{xie2020segmenting} (glass dataset) and PMD \cite{lin2020progressive} (mirror dataset) respectively.

\subsubsection{Visualization on Trans10k dataset}
\label{ssec:trans10k}
\begin{figure*}[!t]
\centering
\includegraphics[scale=1.1]{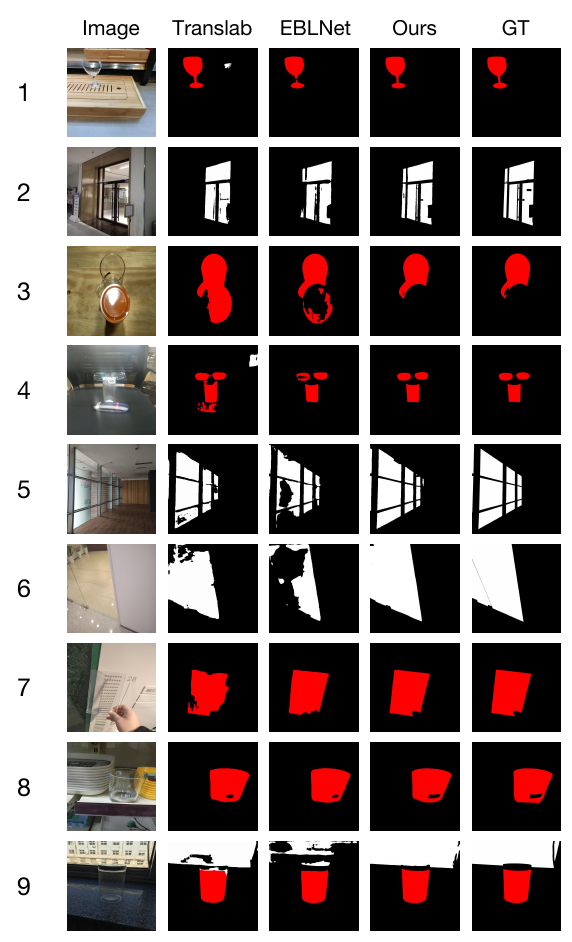}
\caption{Qualitative comparison between the proposed RFENet and two state-of-the-art methods (EBLNet and Translab) on Trans10k dataset. Compared with other methods, our model can better handle various complex scenarios.}
\label{trans_supp}
\end{figure*}

As shown in Figure \ref{trans_supp}, we exhibit the qualitative results of our proposed RFENet and two recent strong competitors, \textit{i.e.}, EBLNet \cite{he2021enhanced} and Translab \cite{xie2020segmenting}. 
Specifically, in the first and second rows, existing methods failed to detect those tiny edges. 
And in the third and fourth rows, there are some highlighted regions where the existing methods failed to distinguish. 
The fifth to sixth rows show some challenging images with huge glass areas. 
The seventh to eighth rows show images with occlusion in front of their glass region, while the existing methods mistake the occlusion as part of the glass. 
The last row shows the input with different types of glass overlapping, in which the overlapped areas are more difficult to distinguish.
On the contrary, our method segments the glass region correctly in all these challenging cases, which demonstrates the superiority of our method over the state-of-the-arts.

\subsubsection{Visualization on PMD dataset}
\label{ssec:pmd}
\begin{figure*}[!t]
\centering
\includegraphics[width=0.9\textwidth]{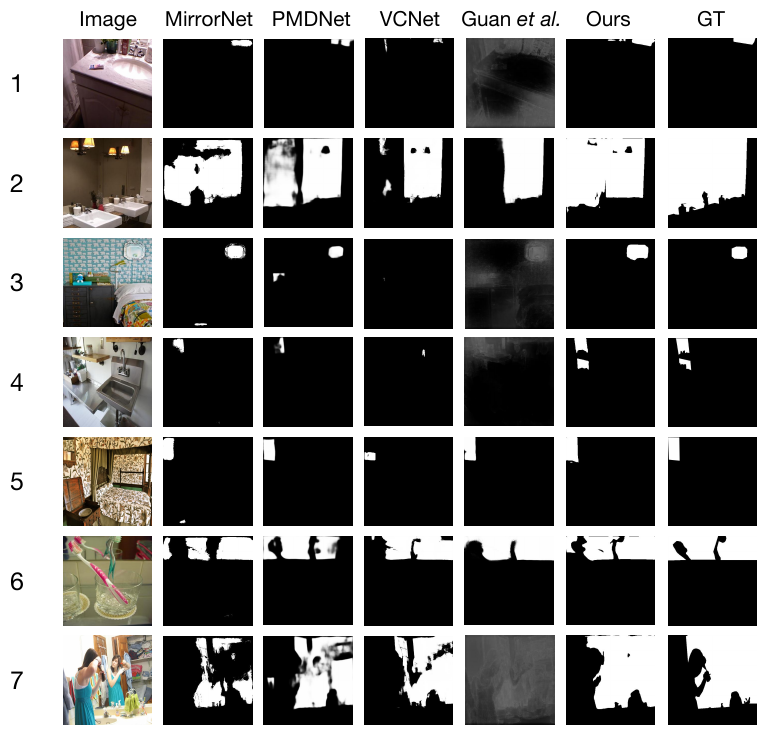}
\caption{Qualitative comparison between the proposed RFENet and four state-of-the-art methods 
(MirrorNet, PMDNet, VCNet and Guan \textit{et al.}) on PMD mirror dataset. 
The promising results show that our method also has transferability on mirror data.}
\label{pmd_supp}
\end{figure*}

We then show the visualized results on PMD \cite{lin2020progressive} to verify the transferability of our method on the mirror dataset. We compare our RFENet with four state-of-the-art methods \cite{tan2022mirror,guan2022learning,lin2020progressive,yang2019my}, as shown in Figure \ref{pmd_supp}. 
In the first row, the mirror is too tiny to be identified, where most of the existing methods failed to detect the mirror. 
The second row shows an image with complex illumination. Existing methods tend to segment only parts of the mirror region. 
The third to fifth rows show images with complex scenarios, where some existing methods failed to detect the less salient mirrors.
In the last two rows, the reflected object overlaps with the mirror in a large area, which also leads to the inferior performance of the existing methods. In contrast, our model still achieves promising results.

In conclusion, all these visualizations demonstrate our model’s transferability even though it is primarily designed for glass-like objects. This indicates our design might also be inspiring for mirror segmentation.

\subsection{Additional Details}
\label{sec:additional details}

\subsubsection{Inference time}
\label{ssec:inference time}
As shown in Table \ref{tab:inference_time}, compared with
the SOTA method, the RFENet with our two key contributions (row2) already achieves a better result with a comparable computation cost, demonstrating the high efficiency of our key idea (i.e., the reciprocal feature evolution).

\subsubsection{Stability analysis}
\label{ssec:stability analysis}
We further run our RFENet on Trans10k datasets 3 times. As shown in Table \ref{tab:std}, the standard deviations on three main metrics are lower enough to illustrate that our model could consistently perform well instead of random coincidence.

\begin{table}[]

\resizebox{\columnwidth}{!}{%
\begin{tabular}{ccccc}
\hline
Methods             & params & fps  & FLOPs & mIoU  \\ \hline
EBLNet              & 111.5M & 12.1 & 304.2 & 90.28 \\ \hline
RFENet(w/o cascade) & 111.9M & 11.9 & 324.7 & 90.86 \\ \hline
RFENet              & 152.6M & 7.3  & 645.4 & 91.25 \\ \hline
\end{tabular}%
}
\caption{The comparison of params, fps and FLOPs with state-of-the-art method EBLNet.}
\label{tab:inference_time}
\end{table}

\begin{table}[]
\centering

\begin{tabular}{cccc}
\hline
Method & mIoU & Acc & mBER \\ \hline
RFENet & 0.03 & 0.05 & 0.04 \\ \hline
\end{tabular}
\caption{The standard deviation on Trans10k by randomly running our RFENet 3 times.}
\label{tab:std}
\end{table}

\clearpage
\section*{Acknowledgements}
This work was supported by the 
National Natural Science Foundation of China (No. 61972157, No. 72192821),
Shanghai Municipal Science and Technology Major Project (2021SHZDZX0102),
Shanghai Science and Technology Commission (21511101200),
Shanghai Sailing Program (22YF1420300, 23YF1410500),
CCF-Tencent Open Research Fund (RAGR20220121),
Young Elite Scientists Sponsorship Program by CAST (2022QNRC001).
\bibliographystyle{named}
\bibliography{ijcai23}
\end{document}